\documentclass[letterpaper]{article}
\usepackage{iccc}
\usepackage{comment}
\usepackage{hyperref}
\usepackage{graphicx}
\usepackage{times}
\usepackage{helvet}
\usepackage{courier}
\usepackage{pgfplots}\pgfplotsset{compat=1.18}
\usepackage{standalone}
\usepackage[super]{nth}
\usepackage{enumitem}
\usepackage{stfloats}
\usepackage{makecell}
\usepackage{array,multirow}
\usepackage{amsmath,amssymb}
\newcommand{\citet}[1]{\citeauthor{#1}\ \shortcite{#1}} 
\newcommand\tstrut{\rule{0pt}{2.0ex}} 
\newcommand\ttstrut{\rule{0pt}{2.8ex}} 

\newcommand\bbstrut{\rule[-1.5ex]{0pt}{0pt}}

\newcommand\exstyle{\sffamily\footnotesize}
\let\mc=\multicolumn
\newcommand\expl[1]{\mc{2}{>{\footnotesize\hangindent=1em$\blacktriangleright\,$}p{.95\columnwidth}}{#1\bbstrut}}
\newcommand\st[1]{\mc{2}{l}{\ttstrut\rmfamily\textbf{#1}}}

\title{The truth is no diaper: Human and AI-generated associations to emotional words}
\author{Špela Vintar$^{\ast,\dagger}$ \& Jan Jona Javoršek$^{\dagger}$ \\ $^{\ast}$University of Ljubljana, Slovenia \\ $^{\dagger}$ Jožef Stefan Institute, Ljubljana, Slovenia \\ \{spela.vintar, jona.javorsek\}@ijs.si}

\begin{document} 
\maketitle
\begin{abstract}
\begin{quote}
Human word associations are a well-known method of gaining insight into the internal mental lexicon, but the responses spontaneously offered by human participants to word cues are not always predictable as they may be influenced by personal experience, emotions or individual cognitive styles. The ability to form associative links between seemingly unrelated concepts can be the driving mechanisms of creativity. We perform a comparison of the associative behaviour of humans compared to large language models. More specifically, we explore associations to emotionally loaded words and try to determine whether large language models generate associations in a similar way to humans. We find that the overlap between humans and LLMs is moderate, but also that the associations of LLMs tend to amplify the underlying emotional load of the stimulus, and that they tend to be more predictable and less creative than human ones.
\end{quote}
\end{abstract}

\section{Introduction}

The free association task is a simple method of eliciting spontaneous connections between words from individuals. Given a cue word, such as \textit{ fish}, the participant is asked to respond with one or several words that come to mind, such as \textit{trout}, \textit{sea bass}, \textit{lake}, \textit{fishing\/} or \textit{ fresh}. The technique was first used in psychology by \citet{galton1880imagery} and later by Freud to examine his patients’ thoughts, in particular hesitations and emotional reactions to certain stimuli \cite{freud1913treatment}. By mid-\nth{20} century, the association task was considered a prominent way of accessing mental representation and memory, and was used extensively in cognitive psychology and cognitive science \cite{deese1959influence}. 

More recently, a number of research studies have been dedicated to comparing internal semantic representations (including word associations) to distributional semantics models and, in the past few years, large language models \cite{mandera_17_explaining,nematzadeh2017evaluating,Gunther2019VectorSpaceMO}. While first generation LLMs would generate a lot of erratic associations where the link between the cue and response was obscure at best \cite{brglezHowHumanLikeAre2024}, state-of-the-art models seemingly excel at mimicking almost any linguistic behaviour. Associations lie at the core of many creative endeavours, and - as recent studies show - can significantly improve tasks like humour generation  \cite{tikhonovHumorMechanicsAdvancing}. But challenges remain, as their performance significantly drops in non-major languages \cite{xuanMMLUProXMultilingualBenchmark2025}, and their ability to form associations in a human-like manner remains understudied. 

We focus on associations to emotionally loaded words, with two research questions in mind: 1. Do LLM-generated associations resemble human ones in terms of creativity, type and overlap, and 2. If emotionally loaded words elicit particular types of sentiments in human responses, can we observe the same distribution of positive and negative sentiments in the generated responses? Our experiments are performed for Slovenian, a morphologically rich language from the family of Slavic languages which currently has just over 2 million speakers. 

\section{Related work}

\looseness=1
Before we review other attempts to compare human and computer-generated associations, it is important to present works which categorize the processes in human associative behaviour. A comprehensive explanation is given in \citet{clarkWordAssociationsLinguistic1970}, who distinguishes amongst association eliciting experiments where participants are under major, moderate or no time pressure. If the person is allowed some time, they ``react with rich images, memories, or exotic verbal associations, and these give way to idiosyncratic, often personally revealing responses.'' Under time pressure, associations become more superficial, predictable and closely related to the stimulus. Clark then continues to present two broad categories of responses: \textit{paradigmatic}, where the response falls into the same syntactic category as the stimulus, and \textit{syntagmatic} where it does not; both mechanisms being governed by implicit linguistic rules such as minimum-contrast, feature-deletion and -addition or idiom-completion. A further claim is that paradigmatic associations are much more frequent than syntagmatic ones. 

\citet{fitzpatrick_2007} challenges the belief that human associative behaviour is homogeneous, and her experiments show large variability among adult native speakers in the categories of responses. She introduces a more fine-grained categorization (which we describe in the section Approach) and shows that associative mechanisms reveal an intricate interplay between cues, responses and individual respondents' associative styles.

In recent years, several authors have explored the ability of vector space models to represent conceptual organization. \citet{mandera_17_explaining} performed a detailed evaluation of correlations between human semantic spaces and corpus-based vector representations, whereby for the former they use semantic priming, semantic relatedness judgements and word associations. They find that static neural models outperform traditional count models, and that the window size used in training plays a significant role in the performance of the models. 

In an experiment by \citet{nematzadeh2017evaluating} human word associations were compared to nearest neighbours suggested by word2Vec and GloVe, and they show that overall correlation is low and that static word embeddings fail to capture certain critical aspects of human associations. A similar conclusion was proposed by \citet{Gunther2019VectorSpaceMO}, who emphasize that ``word meanings are acquired through experience''. For this reason, models trained directly on introspective data generally outperform corpus-trained ones \cite{de2016predicting}.

A more recent detailed discussion of the complexity of human associative behaviour and neural modelling is provided by \citet{richie2022free} who also train their GloVe model on English SWOW \cite{dedeyne_swow18} and achieve good prediction results using a variety of asymmetric measures. 

Finally, \cite{lavoratiLLMGeneratedWordAssociation2024} generate the entire SWOW dataset with Mistral and present some properties of LLM-generated associations against human ones. While no detailed comparison of response categories was performed, the LLM generated almost 3 times fewer unique responses than humans, showing that the diversity and variability of human responses is much higher than for those generated by the current models.  

Our own experiments extend the work cited above in that we focus on emotional words and their polarity, we compare the novelty and creativity of human vs. LLM-generated associations and compute average overlap between them. 

\section{Datasets and methods}
We use the recently constructed SWOW-SL dataset as the source of human association norms. SWOW-SL\footnote{\url{http://hdl.handle.net/11356/1980}} was compiled as the Slovenian branch of the Small World of Words project\footnote{\url{https://smallworldofwords.org}} \cite{dedeyne_swow18}, a site which currently collects word associations for 19 languages of the world including Slovenian. The SWOW-SL v1.0 consists of human responses to 1,000 cue words, totalling over 60,000 associations (20,186 unique responses) from 1396 participants. 
Since our focus is on emotionally loaded words, our second important resource is the Slovenian Emotion Dimension and Emotion Association Lexicon SloEmoLex 1.0 \cite{brglez2024slovenian}, an extension of the LiLaH CroSloEng Emotion lexicon created by \citet{ljubesic-etal-2020-lilah}. The lexicon contains Valence, Dominance and Arousal scores for almost 20,000 Slovenian words, of which around 14,000 are also assigned binary variables for Positive and Negative sentiment along with a discrete model of emotion covering anger, anticipation, disgust, fear, joy, sadness, surprise and trust.

Using the score for Valence which is generally considered as a measure of emotional load we select top 20 and lowest 20 Slovenian words, whereby the words must also be included as cues in the SWOW-SL association database. These words represent samples of positively and negatively loaded parts of the vocabulary, and include:

\begin{description}
\item[Positive:] \textit{resnica} (truth), \textit{odlično} (excellent), \textit{zmagovalec} (winner), \textit{srečen} (happy), \textit{ sposoben} (skillful), \textit{zadovoljen} (content), \textit{zabava} (fun), \textit{mir} (peace), \textit{zlat} (gold), \textit{svoboda} (freedom), \textit{zdrav} (healthy), \textit{mama} (mom), \textit{mati} (mother), \textit{dobiček} (profit), \textit{praznik} (holiday), \textit{sprejet} (accepted), \textit{prijazen} (kind), \textit{sestra} (sister), \textit{ženski} (female), \textit{dogovor} (agreement)
\item[Negative:] \textit{napadalec} (attacker), \textit{padec} (fall), \textit{ povzročiti} (inflict), \textit{nevarnost } (danger), \textit{poraz} (defeat), \textit{strel} (shot), \textit{bolnik} (patient), \textit{nevaren} (dangerous), \textit{žr\-t\-ev} (victim), \textit{ bolečina} (pain), \textit{ smrt} (death), \textit{pomanj\-kanje} (poverty), \textit{pri\-tož\-ba} (complaint), \textit{poškodovan} (hurt), \textit{nasprotnik} (opponent), \textit{slabo} (bad), \textit{nasilje} (violence), \textit{zavrniti} (reject), \textit{izdati} (betray), \textit{umreti} (die)
\end{description}

\subsection{Approach}
After selecting the emotional cues, we retrieve associations from the SWOW-SL database, and generate associations using 3 contemporary LLMs: 

\begin{itemize}
\item Llama-3.3: a text-only 70B instruction-tuned model by Meta,
\item GaMS-9B: a new improved and larger model of the GaMS (Generative Model for Slovene) family, based on Google's Gemma 2 family and continually pretrained on Slovene, English and some portion of Croatian, Serbian and Bosnian corpora, developed by CJVT\footnote{Centre for Language Resources and Technologies at the University of Ljubljana.},
\item Claude 3.7 Sonnet: a commercial reasoning model developed by Anthropic and an enhanced version of Sonnet 3.5, number of parameters not public. 
\end{itemize}

\begin{table*}[th!]
    \begin{tabular}[t]{ >{\exstyle}l >{\exstyle}l }
    \mc{1}{c}{\rmfamily\textbf{Cue} [\rmfamily\textbf{Gloss}]}&\mc{1}{c}{\rmfamily\textbf{Response} [\rmfamily\textbf{Gloss}]}\\
    \hline\tstrut
    \textit{resnica} [truth] & \textit{ni plenica} [no diaper] \\
    \expl{`the truth is no  diaper', words rhyme}\\
    \textit{resnica} [truth] & \textit{skazica} [turns.out]\\
    \expl{`truth turns out': rhyming allusion to a common expression \textit{pravica -- skazica} (what is just is rewarded in the end)}\\
    \textit{odlično} [excellent] & \textit{oreščki} [nuts]\\
    \expl{refering to \textit{Odlično} nuts and seeds brand in Slovenia}\\
    \textit{odlično} [excellent] & \textit{lari fari} [blah blah]\\
    \textit{srečen} [happy] & \textit{vrvica} [string] \\
    \expl{probably a reference to an old Slovenian movie \textit{Sreča na vrvici} [Happiness on a leash] about a boy and his dog}\\
    \textit{srečen} [happy] & \textit{lonec} [pot] \\
    \expl{probably a reference to the rhyme \textit{srečen konec, počen lonec} [happy end -- broken pot]}\\
    \textit{zmagovalec} [winner] & \textit{the winner takes it all (abba)} \\
    \expl{reference to an Abba song}\\
    \textit{sposoben} [capable] & \textit{sanjati} [of.dreaming]  \\
    \textit{mir} [peace] & \textit{megla} [fog]\\
    \textit{mir} [peace] & \textit{golobica} [dove] \\
    \expl{reference to the dove as a symbol of peace}\\
    \textit{mama} [mom] & \makecell{ \textit{stara mama k štrudl peče} \\ {[granny baking strudel]} } \\
    \textit{mama} [mom] & \textit{rdeča} [red]\\
    \textit{mama} [mom] & \textit{neprespana} [sleepless] \\
    \textit{mama} [mom] & \textit{mari} [Mari] \\
    \expl{a female name}\\
    \textit{mama} [mom] & \textit{luč} [light] \\
    \textit{strel} [shot] & \textit{sarajevo} [Sarajevo] \\
    \expl{probably a reference to the Bosnian capital under siege}\\
    \textit{nasilje} [violence] & \textit{črna} [black] \\
  \end{tabular}
    \qquad
  \begin{tabular}[t]{ >{\exstyle}l >{\exstyle}l }
    \mc{1}{c}{\rmfamily\textbf{Cue} [\rmfamily\textbf{Gloss}]}&\mc{1}{c}{\rmfamily\textbf{Response} [\rmfamily\textbf{Gloss}]}\\
    \hline\tstrut
    \textit{smrt} [death] & \textit{ja} [yes] \\
    \textit{smrt} [death] & \textit{deževna trata} [rainy meadow]\\
    \expl{possibly personal reference}\\
    \textit{smrt} [death] & \makecell{ \textit{nena ga več vidiš} \\ {[Nena can you see it yet]} } \\
    \expl{a personal comment (in Croatian)}\\
    \textit{smrt} [death] & \textit{komot} [easily] \\
    \textit{smrt} [death] & \makecell{ \textit{zakaj je banana rumena} \\ {[why are bananas yellow]} } \\
    \expl{an absurdist comment}\\
    \textit{smrt} [death] & \textit{svet je mejhn} [the world is small]\\
    \noalign{\medskip}
    \st{Llama}\\
    \hline\tstrut
    \textit{mir} [peace] & \textit{miran} [`Miran'/peaceful] \\
    \expl{a name, or interference with Croatian for `peaceful'}\\
    \textit{svoboda} [freedom] & \textit{avstralija} [Australia] \\
    \textit{smrt} [death ] & \textit{pepelka} [Cinderella]\\
    \textit{zavrniti} [reject] & \textit{opoziv} [?] \\
    \expl{a non-existing word resembling `opposition' and `appeal'}\\
    \st{claude}\\
    \hline\tstrut
    \textit{poraz} [defeat] & \textit{nezmaga} [non-victory] \\
    \expl{a made-up word}\\
    \textit{žrtev} [victim] & \textit{trpin} [sufferer] \\
    \expl{a rare word}\\
    \st{GaMS}\\
    \hline\tstrut
    \textit{zlat} [golden] & \textit{goldman} [`goldman'] \\
    \expl{possibly interference from English}\\
    \textit{ženski} [female] & \textit{spolov} [`of.gender']\\
    \end{tabular}
    \caption{\label{creative_responses}Examples of unexpected, novel or creative responses from human and LLM contributions}
\end{table*}

The prompt to generate association is zero-shot: ``What does the word X remind you of?'', with the specific instruction to all models to provide only 3 associations to each cue. Once the models provided their responses we perform several sets of analyses: firstly, we examine the association inventories and look for unexpected, novel or creative responses; secondly, we compute overlap between all 4 versions (human vs llama/gams/claude); and thirdly, we label associations with Valence scores and examine the average emotional polarity of each setting. 

\subsubsection{Creative responses.}
For the purposes of this manual analysis, we loosely rely on the categories of responses as defined by Fitzpatrick \shortcite{fitzpatrick_2007}:
\begin{itemize}
    \item \textbf{Meaning-based association}: \textit{x} means the same as \textit{y}, \textit{x} and \textit{y} come from the same lexical set, \textit{x} and \textit{y} have some other conceptual link
    \item \textbf{Position-based association}: \textit{y} follows or precedes \textit{x} directly or with words between them
    \item \textbf{Form-based association}: \textit{y} is \textit{x} plus or minus affix, \textit{y} looks or sounds similar to \textit{x}
    \item \textbf{Erratic}: \textit{y} has no decipherable link to \textit{x} or no response given
\end{itemize}

The fourth category is the one potentially containing the most creative responses, although other categories may also involve mechanisms which evoke unexpected responses in humans based on rhyme, memories or personal idiosyncrasies. We inspect human and generated associations looking for responses which go beyond typical conceptual or position-based links between cue and response.

\medskip

\noindent\textbf{Human.} %
A manual analysis of human associations shows the great variability, individuality and creativity in the responses, where the relations between cues and responses range from wordplay, intertextual or intercultural references up to completely obscure associations where the connection evades explanation. Some examples with attempted gloss translations and explanations are listed in Table \ref{creative_responses}.

\medskip

\noindent\textbf{LLMs.}%
As opposed to the richness of the human associative space, LLMs produce associations which are far less original and may only occasionally depart from the expected meaning- or position-based responses, so the number of examples in Table \ref{creative_responses} is rather small. In addition, LLM-generated associations typically follow the part-of-speech category of the cue or remain within the pool of morphologically related words (eg. \textit{nevaren} [dangerous]: \textit{tvegan} [risky], \textit{hazarden} [hazardous], \textit{nevarnost} [danger], \textit{varovanje} [protection]) as opposed to human responses which are truly associative in that they are in part motivated by the underlying emotion (\textit{nevaren} [dangerous]: \textit{moški} [man], \textit{pes} [dog], \textit{človek} [(hu)man]). The claim that paradigmatic associations are much more frequent than syntagmatic ones \cite{clarkWordAssociationsLinguistic1970} is thus even truer for LLMs. 


\subsection{Overlap}
We compute average overlap by intersecting sets of responses for each cue in each setting (Table \ref{overlap}). The model corresponding most closely to human associations is the presumably largest, Claude 3.7 Sonnet, followed by Llama and GaMS respectively. As for correspondences between models, the highest match is between Claude and the Slovenian GaMS, which perhaps means that additional target language data improves performance on the association task; an assumption otherwise not proven by the Human-GaMS overlap. 

\begin{table}[hbt!]
\centering
\begin{tabular}{lccc}
\hline
\tstrut\textbf{} & \textbf{Human} & \textbf{Llama} & \textbf{GaMS}\\
\hline
Llama & 18.33 & / & 12.50 \tstrut\\
GaMS & 14.17 & 12.50 & / \\
Claude & 23.33 & 14.17 & 17.50  \\
\hline
\end{tabular}
\caption{\label{overlap} Overlap between human and models' associations}
\end{table}

\subsection{Sentiment analysis}
\looseness=-1
In the subsequent analysis of the sentiment of associations we label each response with its Valence score (between 0 and 1, 0 signifying extremely negative, $0.50$ neutral and 1 extremely positive) and the binary Positive and Negative scores. We analyse the 20 positive and 20 negative cues separately, by computing average scores per cue and each of the two groups.

\begin{table}[hbt!]
\centering
\setlength{\belowcaptionskip}{-8pt}
\begin{tabular}{lccc}
\hline
\tstrut\textbf{} & \textbf{Valence} & \textbf{Positive} & \textbf{Negative}\\
\hline
Human & 0.73 & 0.53	& 0.11 \tstrut\\
Llama & 0.78 & 0.41 & 0.05 \\
Claude & 0.78 & 0.72 & 0.08  \\
GaMS & 0.71 & 0.53 & 0.15 \\
\hline
\end{tabular}
\caption{\label{positive_sentiment} Average sentiment scores for positive cues}
\end{table}

\begin{figure}[ht]
    \centering
    \setlength{\belowcaptionskip}{-8pt}
    \begin{tikzpicture}
  \begin{axis}[
      ybar,
      enlargelimits=0.65,
      legend style={at={(0.5,-0.15)},
        anchor=north,legend columns=-1},
      ylabel={Average valence of responses},
      symbolic x coords={Positive cue,Negative cue},
      xtick=data,
      bar width = 16pt,
      nodes near coords,
      nodes near coords align={vertical},
      nodes near coords style={ font=\footnotesize }
    ]
    \addplot coordinates {(Positive cue, 0.73) (Negative cue, 0.37)};
    \addplot coordinates {(Positive cue, 0.78) (Negative cue, 0.29)}; 
    \addplot coordinates {(Positive cue, 0.78) (Negative cue, 0.29)};
    \addplot coordinates {(Positive cue, 0.71) (Negative cue, 0.25)};
    \legend{Human, Llama, claude, GaMS}
    \end{axis}
  \end{tikzpicture}
    \caption{Valence for positive and negative cues}
    \label{fig:valence}
\end{figure}
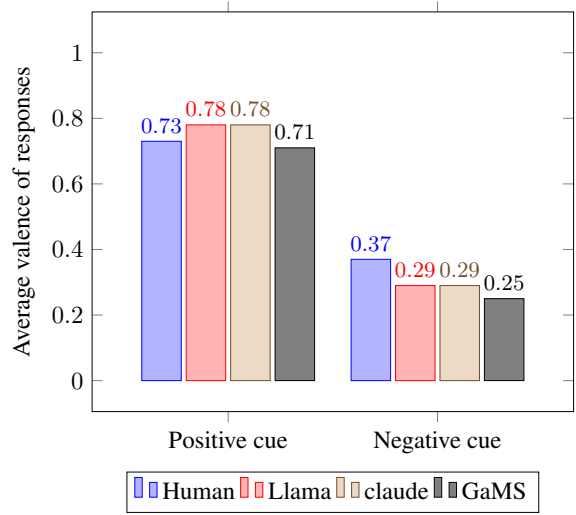

\begin{table}[hbt!]
\centering
\begin{tabular}{lccc}
\hline
\tstrut\textbf{} & \textbf{Valence} & \textbf{Positive} & \textbf{Negative}\\
\hline
Human & 0.37 & 0.07	& 0.47 \tstrut\\
Llama & 0.29 & 0.04 & 0.58 \\
Claude & 0.29 & 0.03 & 0.58  \\
GaMS & 0.25 & 0.07 & 0.58 \\
\hline
\end{tabular}
\caption{\label{negative_sentiment} Average sentiment scores for negative cues}
\end{table}
Tables \ref{positive_sentiment} and \ref{negative_sentiment} show the average sentiment scores of responses to emotional words for positive and negetive cues. As explained above, valence scores higher than $0.50$ indicate positive sentiment, hence all models generate predominantly responses with a positive sentiment to positive cues and vice versa, with Llama and Claude even surpassing humans in both extremes. The columns Positive and Negative record the average number of responses in those two categories, whereby these values are binary in SloEmoLex (a word can be either positive or not), and not all words have this value assigned. Results are most conclusive for negative cues, where it seems that all models generate even more negative associations than humans. (See Figure \ref{fig:valence}.)

\section{Discussion and conclusions}
The experiments described show that LLMs are able to generate association-like responses to a zero-shot prompt, even in a small language like Slovene. A fair portion of these responses overlapped with human association norms, with the largest model (Claude 3.7 Sonnet) achieving the highest overlap. It should be noted here that seemingly small numbers do not necessarily mean poor results, as human-human overlap typically does not exceed 40 \%. 

In line with our own previous work \cite{brglezHowHumanLikeAre2024}, LLM-generated associations generally follow the paradigmatic pattern more often than the syntagmatic one, frequently listing synonyms or near synonyms, and rarely departing from the part-of-speech of the cue. While the present study did not focus specifically on the comparison of response categories, a manual inspection of the results indicates slight variations between models in their typical behaviour. It would appear that LLMs come in different flavours, or exhibit distinct and consistent personal traits which affect - among other things - their linguistic behaviour. This should not surprise us as recent studies within the newly emerging field of machine psychology indeed explore personality traits of LLMs \cite{tommasoLLMsPersonalitiesInconsistencies2024,heInvestigatingImpactLLM2025}.

When human and LLM responses are compared in terms of creativity or unexpectedness, humans produce a much richer and more varied mix of responses. We are aware of the fact that the creativity of LLMs could be activated using a different prompt, possibly including a variety of human responses or explicitly requesting creativity. We deliberately refrained from prompt engineering to retain the similarity between the association task for humans and LLMs.

The sentiment analysis tentatively shows that the models tend to follow the sentiment of the cue and only rarely depart from it in a different direction, rather the responses reinforce and often amplify the original sentiment. This finding is in line with the observation of \citet{lavoratiLLMGeneratedWordAssociation2024} about LLM bias in generated associations: if human responses to \textit{man} and \textit{woman} contained traces of gender stereotypes (\textit{woman: sex, beauty; man: strong}, Mistral's responses were overtly stereotypical (\textit{woman: makeup, hair, fashion, beauty; man: job, career, computer, work}). One reason for the exaggeration of the original sentiment might be that the models almost never generate the antonym as response, while this is a strong association mechanism in humans. An antonym typically has the reverse sentiment than the cue, so this might explain why the average Valence scores for human responses lie more towards the middle of the range as compared to LLMs.

Our findings contradict the observations of \citet{guoHowCloseChatGPT2023} who claim that ChatGPT's responses exhibit less bias and more objectivity than human answers to a collection of questions. This may be due to the fact that question answering, the task that instruction-tuned LLMs have been specifically trained for, triggers different behaviour than the association task where the model is prompted to explore its cognitive and experiential space (\textit{"What does the word X remind you of?"}). 

Even if the (lack of) diversity and variability in LLM-generated responses may say little about their (lack of) creativity, we believe that the tendency to produce outputs which follow the grammatical category, sentiment and semantic domain of the cue captures but a fraction of human associative behaviour, and that departures from the expected -- which used to be quite common in 1st generation LLMs -- now seem to be distinctive traits of human associations. 

In future experiments it would make sense to perform a more detailed and fine-grained categorization of responses in order to see how models differ in their association patterns, a phenomenon we have observed but not yet evaluated. Also, the sample of 40 cues is too small to allow for any generalization of the above observations.   

%
%
%

\section{Author Contributions}

Špela Vintar was in charge of planning the study, conducted parts of the analysis and wrote a significant part of the manuscript. Jan Jona Javoršek participated in the analysis, evaluation and interpretation of the results, and contributed to all versions of the manuscript.

\section{Acknowledgments}

This contribution was partly funded by ARIS grants P6-0215 Slovene Language - Basic, Contrastive, and Applied Studies, GC-0002 Large Language Models for Digital Humanities and I0-0005 The JSI infrastructure program.






\bibliographystyle{iccc}
\bibliography{iccc}

\end{document}